\documentclass{article}
\usepackage{ijcai22}

\usepackage{xr-hyper}
\makeatletter
\newcommand*{\addFileDependency}[1]{%
  \typeout{(#1)}%
  \@addtofilelist{#1}%
  \IfFileExists{#1}{}{\typeout{No file #1.}}%
}
\makeatother

\usepackage{times}

\usepackage{soul}
\usepackage{url}
\usepackage[hidelinks]{hyperref}
\usepackage[utf8]{inputenc}
\usepackage[small]{caption}
\usepackage{graphicx}
\usepackage{amsmath}
\usepackage{amssymb}

\usepackage{booktabs}
\usepackage{siunitx}
\usepackage{etoolbox}
\renewrobustcmd{\bfseries}{\fontseries{b}\selectfont}
\renewrobustcmd{\boldmath}{}
\usepackage{multirow}

\usepackage{pifont}
\usepackage[dvipsnames]{xcolor}
\usepackage{xspace}
\usepackage{appendix}
\AtBeginEnvironment{appendices}{\crefalias{section}{appendix}}

\usepackage[capitalise]{cleveref}
\usepackage{subfiles}

\usepackage{caption}
\usepackage{subcaption}

\usepackage{multibib}
\newcites{SM}{Supplementary Material References}

\newcommand{\tick}{\ding{52}}

\newcommand{\cross}{--}

\newcommand{\methodname}{ChimeraMix\xspace}
\newcommand{\methodnamegrid}{ChimeraMix+Grid\xspace}
\newcommand{\methodnameseg}{ChimeraMix+Seg\xspace}
\newcommand{\textcite}[1]{\citeauthor{#1}\xspace\shortcite{#1}}

\newcommand{\cifairX}{ciFAIR-10\xspace}
\newcommand{\cifairC}{ciFAIR-100\xspace}
\newcommand{\stl}{STL-10\xspace}

\usepackage[nolist,nohyperlinks]{acronym}
\begin{acronym}
\acro{FID}{Fréchet Inception Distance}
\acro{MDE}{Minimum Distortion Embedding}
\acro{GLICO}{Generative Latent Implicit Conditional Optimization}
\end{acronym}

\title{\methodname: Image Classification on Small Datasets via Masked Feature Mixing}

\author{
Christoph Reinders\footnote{Equal contribution}\and
Frederik Schubert$^*$\And
Bodo Rosenhahn\\
\affiliations
Institute for Information Processing, Leibniz University Hannover
\emails
\{reinders,schubert,rosenhahn\}@tnt.uni-hannover.de
}

\begin{document}

\maketitle

\begin{abstract}
Deep convolutional neural networks require large amounts of labeled data samples. 
For many real-world applications, this is a major limitation which is commonly treated by augmentation methods. 
In this work, we address the problem of learning deep neural networks on small datasets. 
Our proposed architecture called \emph{\methodname} learns a data augmentation by generating compositions of instances. 
The generative model encodes images in pairs, combines the features guided by a mask, and creates new samples.
For evaluation, all methods are trained from scratch without any additional data.
Several experiments on benchmark datasets, e.g., \cifairX, \stl, and \cifairC, demonstrate the superior performance of \emph{\methodname} compared to current state-of-the-art methods for classification on small datasets. 
Code is available at \\\url{https://github.com/creinders/ChimeraMix}.
\end{abstract}

\section{Introduction}
\label{sec:introduction}

Large-scale datasets contribute significantly to the success of deep neural networks in computer vision and machine learning in recent years.
The collection of massive amounts of labeled data samples, however, is very time-consuming and expensive. 
Less explored is the research direction of applying deep learning algorithms on small data problems.
These small data problems are common in the real world.
In many applications, there is not much data available or cannot be used due to legal reasons \cite{renardVariabilityReproducibilityDeep2020}.

When learning with a limited amount of data, most research focuses on transfer learning, self-supervised learning, and few-shot learning techniques. 
Transfer \cite{neyshaburWhatBeingTransferred2020} and self-supervised methods \cite{assranSemiSupervisedLearningVisual2021} generate a representation on a large source dataset and transfer the knowledge to the target domain where the model can be fine-tuned. 
Similarly, few-shot learning methods \cite{kolesnikovBigTransferBiT2020} are trained on a base set to generalize to a novel set given a small number of support examples. 
All these approaches, however, require a large source dataset of annotated data samples and the source domain needs to be close to the target domain.
Additionally, different input sensors (e.g., hyperspectral camera or depth sensor) or copyright conditions often prevent the use of readily available datasets.

In this work, we address the challenging small data problem by presenting a novel feature mixing architecture for generating images. 
Our method makes use of the fact that the label of the classification task is invariant under the composition of object instances.
Thus, given multiple instances of a class our generative model is able to compose new samples guided by a mask. 
We present two methods to generate the mixing masks, one based on a grid of rectangular patches and one that uses a segmentation algorithm.

\begin{figure}[t]
	\centering
	\includegraphics[width=\linewidth]{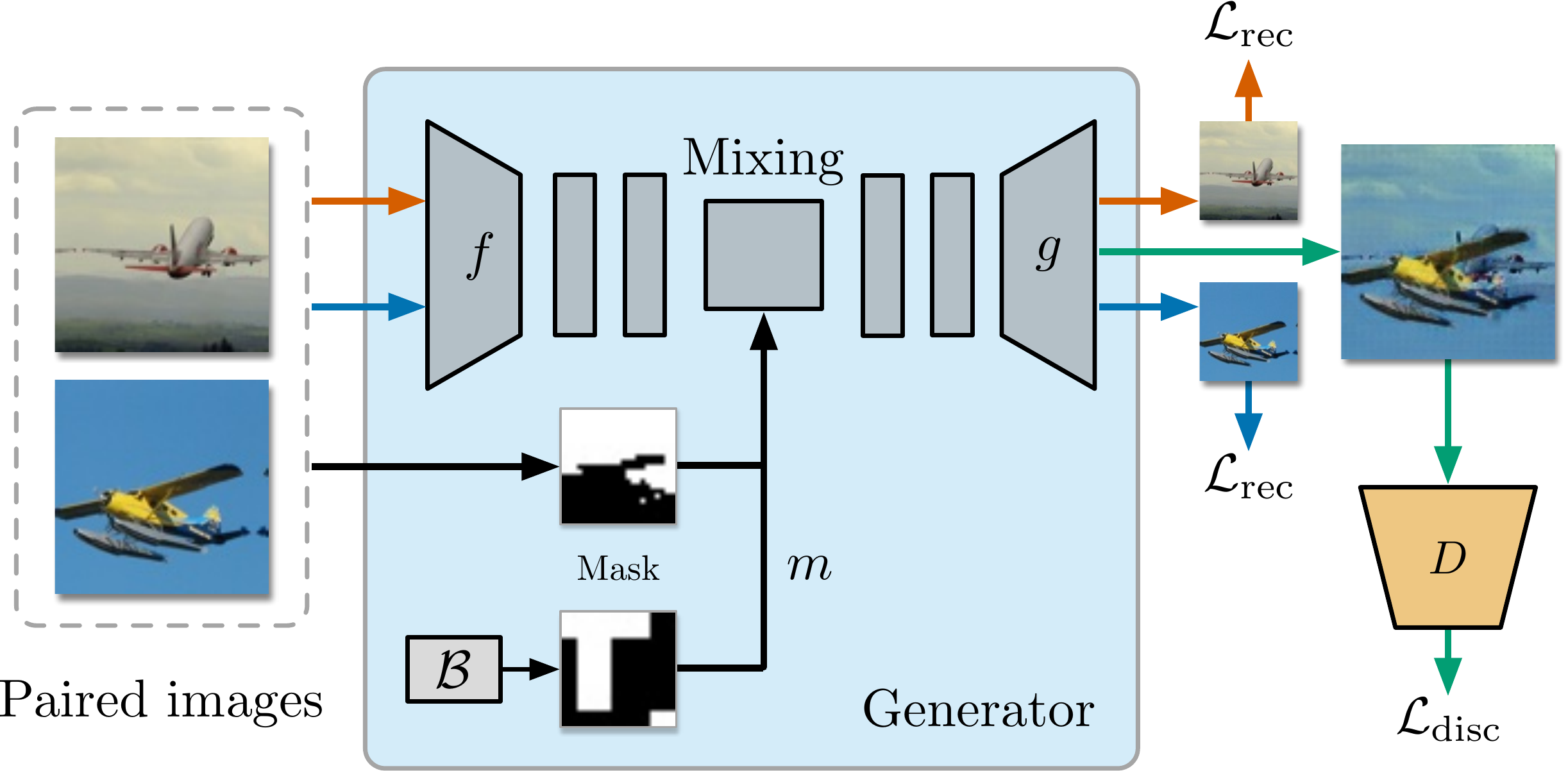}	
	\caption{
		\textbf{Overview of \methodname} – Two images of the same class are passed through a feature extractor $f$.
		Their features are mixed according to a mask $m$ and the generator-discriminator architecture learns to generate new samples. 
		Additionally, a reconstruction loss $\mathcal{L}_{\text{rec}}$ improves the sample quality and stabilizes the training process of \methodname. 
		The mask is either sampled from a binomial distribution $\mathcal{B}$ or computed using a segmentation algorithm.
	}
	\label{fig:architecture}
\end{figure}

\begin{figure}[!ht]
     \centering
     \includegraphics[width=\linewidth]{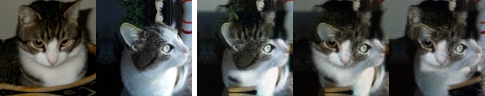}
     \vspace{0.25cm}
     \includegraphics[width=\linewidth]{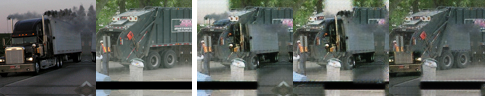}
     \includegraphics[width=\linewidth]{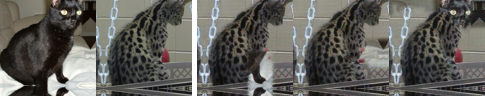}
     \includegraphics[width=\linewidth]{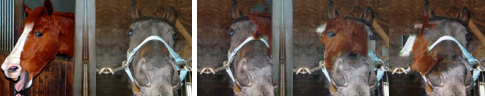}
    \caption{Examples of generated images by \methodnamegrid (top) and \methodnameseg (bottom) on \stl. \methodname combines image pairs (first two columns), mixes the features guided by a sampled mask and generates new image compositions (last three columns).}
    \label{fig:stl10_samples}
\end{figure}

The generative process and training of our method \emph{\methodname} is outlined in \cref{fig:architecture}.
Overall, we construct our method as a generative-adversarial architecture.
The masks are used to mix the features at different locations in the images.
Subsequently, the generator network decodes and refines the features to create new compositions (examples are shown in \cref{fig:stl10_samples}).
We evaluate our method on benchmarks and raise the state-of-the-art accuracy in the small data setting on \cifairX, \stl, and \cifairC.
Additionally, we provide in-depth analyses to demonstrate the impact of the generator and combine our method with different automatic augmentation methods to improve the accuracy even further.

To summarize, our \textbf{contributions} are:
\begin{itemize}
    \item We propose a novel generative approach for addressing the small data image classification task.
    \item Our generator introduces a feature mixing architecture. Guided by a mask, the generator learns to combine images and create new compositions of the instances.
    \item Experiments on benchmark datasets demonstrate that \methodname outperforms current state-of-the-art methods in small data image classification.
    \item Our experiments show that \methodname can be combined with other augmentation methods to improve their performance even further.
\end{itemize}

\begin{table*}[ht]
\centering
\label{tab:results_cifair10_stl10}
\begin{tabular}{llS[table-format=2.2(2)]S[table-format=2.2(2)]S[table-format=2.2(2)]S[table-format=2.2(2)]S[table-format=2.2(2)]S[table-format=2.2(2)]}
\toprule
\multicolumn{2}{l}{Samples per Class} & {5} & {10} & {20} & {30} & {50} & {100} \\
{Dataset} & {Method} & {} & {} & {} & {} & {} & {} \\
\midrule
\multirow[c]{9}{*}{ciFAIR-10} & Baseline & \SI{31.37\pm3.28}{} & \SI{38.09\pm1.34}{} & \SI{47.50\pm2.09}{} & \SI{53.19\pm0.60}{} & \SI{58.84\pm0.82}{} & \SI{70.34\pm1.17}{} \\
 & Cutout & \SI{28.88\pm2.84}{} & \SI{37.33\pm1.03}{} & \SI{47.55\pm2.06}{} & \SI{53.39\pm1.32}{} & \SI{61.17\pm1.03}{} & \SI{72.14\pm1.10}{} \\
 & Random Erasing & \SI{28.91\pm2.64}{} & \SI{37.13\pm0.61}{} & \SI{47.20\pm2.32}{} & \SI{53.11\pm1.65}{} & \SI{60.34\pm0.35}{} & \SI{72.00\pm0.71}{} \\
 & Cosine & \SI{31.45\pm3.22}{} & \SI{37.88\pm1.24}{} & \SI{46.69\pm1.38}{} & \SI{52.16\pm0.72}{} & \SI{59.24\pm1.60}{} & \SI{70.18\pm1.32}{} \\
 & MixUp & \SI{33.41\pm2.70}{} & \SI{43.03\pm1.21}{} & \SI{53.09\pm1.00}{} & \SI{59.47\pm1.10}{} & \SI{66.16\pm0.78}{} & \SI{74.23\pm0.35}{} \\
 & Scattering & \SI{30.50\pm3.87}{} & \SI{37.28\pm1.87}{} & \SI{45.65\pm1.45}{} & \SI{50.47\pm1.19}{} & \SI{54.30\pm0.95}{} & \SI{61.51\pm0.79}{} \\
 & GLICO & \SI{31.91\pm2.41}{} & \SI{42.02\pm0.87}{} & \SI{51.61\pm1.23}{} & \SI{59.03\pm0.70}{} & \SI{65.00\pm1.24}{} & \SI{73.96\pm0.81}{} \\
 & ChimeraMix+Grid & \SI{36.94\pm2.63}{} & \SI{45.57\pm2.11}{} & \SI{53.67\pm2.84}{} & \SI{59.66\pm1.35}{} & \SI{65.42\pm0.83}{} & \SI{73.76\pm0.30}{} \\
 & ChimeraMix+Seg & \bfseries \SI{37.31\pm2.57}{} & \bfseries \SI{47.60\pm1.81}{} & \bfseries \SI{56.21\pm1.77}{} & \bfseries \SI{60.92\pm0.62}{} & \bfseries \SI{67.30\pm1.21}{} & \bfseries \SI{74.96\pm0.21}{} \\
\midrule
\multirow[c]{8}{*}{STL-10} & Baseline & \SI{27.61\pm0.90}{} & \SI{31.93\pm1.68}{} & \SI{36.50\pm0.94}{} & \SI{39.95\pm1.26}{} & \SI{44.82\pm0.48}{} & \SI{53.51\pm1.65}{} \\
 & Cutout & \SI{28.05\pm1.73}{} & \SI{31.45\pm2.46}{} & \SI{37.68\pm1.30}{} & \SI{40.69\pm1.13}{} & \SI{45.63\pm1.19}{} & \SI{54.32\pm1.01}{} \\
 & Random Erasing & \SI{27.87\pm1.36}{} & \SI{31.32\pm0.48}{} & \SI{36.91\pm1.45}{} & \SI{40.66\pm0.84}{} & \SI{45.93\pm1.10}{} & \SI{53.31\pm1.52}{} \\
 & Cosine & \SI{25.97\pm0.93}{} & \SI{30.37\pm1.34}{} & \SI{35.51\pm0.95}{} & \SI{40.05\pm1.01}{} & \SI{45.51\pm1.23}{} & \SI{53.01\pm1.09}{} \\
 & MixUp & \SI{30.06\pm1.80}{} & \SI{35.63\pm0.85}{} & \SI{42.44\pm1.85}{} & \SI{45.00\pm2.71}{} & \SI{49.03\pm1.34}{} & \SI{54.38\pm2.11}{} \\
 & GLICO & \SI{26.97\pm0.98}{} & \SI{33.02\pm1.07}{} & \SI{37.88\pm1.22}{} & \SI{42.66\pm0.66}{} & \SI{48.40\pm0.72}{} & \SI{54.82\pm1.94}{} \\
 & ChimeraMix+Grid & \bfseries \SI{32.18\pm0.90}{} & \SI{37.01\pm0.84}{} & \SI{43.19\pm1.03}{} & \SI{48.93\pm1.34}{} & \SI{52.81\pm1.45}{} & \SI{60.04\pm0.27}{} \\
 & ChimeraMix+Seg & \SI{31.37\pm1.72}{} & \bfseries \SI{37.05\pm1.09}{} & \bfseries \SI{44.74\pm0.60}{} & \bfseries \SI{49.58\pm0.49}{} & \bfseries \SI{55.06\pm1.11}{} & \bfseries \SI{60.44\pm0.71}{} \\
\midrule
\multirow[c]{10}{*}{ciFAIR-100} & Baseline & \SI{18.78\pm0.79}{} & \SI{24.53\pm0.28}{} & \SI{39.27\pm0.31}{} & \SI{45.99\pm0.32}{} & \SI{53.40\pm0.36}{} & \SI{61.81\pm0.41}{} \\
 & Cutout & \SI{19.25\pm0.52}{} & \SI{27.77\pm0.39}{} & \SI{40.72\pm0.68}{} & \SI{47.78\pm0.39}{} & \SI{55.13\pm0.30}{} & \bfseries \SI{63.26\pm0.62}{} \\
 & Random Erasing & \SI{18.35\pm0.37}{} & \SI{26.09\pm0.74}{} & \SI{38.83\pm1.01}{} & \SI{46.14\pm0.38}{} & \SI{54.26\pm0.08}{} & \SI{63.24\pm0.50}{} \\
 & Cosine & \SI{18.04\pm0.87}{} & \SI{23.72\pm0.35}{} & \SI{38.84\pm0.73}{} & \SI{45.83\pm0.43}{} & \SI{53.32\pm0.11}{} & \SI{61.50\pm0.46}{} \\
 & MixUp & \SI{20.63\pm0.16}{} & \SI{31.03\pm0.54}{} & \SI{41.58\pm0.40}{} & \SI{47.88\pm0.45}{} & \SI{54.87\pm0.20}{} & \SI{62.49\pm0.52}{} \\
 & Scattering & \SI{12.67\pm0.40}{} & \SI{18.25\pm0.56}{} & \SI{26.37\pm0.63}{} & \SI{31.51\pm0.28}{} & \SI{36.49\pm0.42}{} & \SI{48.18\pm0.33}{} \\
 & GLICO & \SI{19.32\pm0.39}{} & \SI{28.49\pm0.60}{} & \SI{40.45\pm0.30}{} & \SI{45.90\pm0.77}{} & \SI{53.53\pm0.19}{} & \SI{60.68\pm0.50}{} \\
 & SuperMix & \SI{19.23\pm0.45}{} & \SI{26.78\pm0.20}{} & \SI{38.47\pm0.83}{} & \SI{44.69\pm0.63}{} & \SI{53.07\pm0.13}{} & \SI{62.63\pm0.30}{} \\
 & ChimeraMix+Grid & \SI{20.24\pm0.12}{} & \SI{31.62\pm0.82}{} & \SI{41.80\pm0.52}{} & \SI{48.10\pm0.71}{} & \SI{54.67\pm1.01}{} & \SI{62.13\pm0.27}{} \\
 & ChimeraMix+Seg & \bfseries \SI{21.09\pm0.47}{} & \bfseries \SI{32.72\pm0.60}{} & \bfseries \SI{43.23\pm0.38}{} & \bfseries \SI{48.83\pm0.72}{} & \bfseries \SI{55.79\pm0.21}{} & \SI{62.96\pm0.77}{} \\
\bottomrule
\end{tabular}
\caption{Test accuracy on \cifairX, \stl, and \cifairC.
Cutout \protect\cite{devriesImprovedRegularizationConvolutional2017}, Random Erasing \protect\cite{zhongRandomErasingData2020}, Cosine \protect\cite{barzDeepLearningSmall2020}, MixUp \protect\cite{zhangMixupEmpiricalRisk2018}, Scattering \protect\cite{gauthierParametricScatteringNetworks2021}, GLICO \protect\cite{azuriGenerativeLatentImplicit2021}, and SuperMix \protect\cite{daboueiSuperMixSupervisingMixing2021} are state-of-the-art methods for small data image classification. A standard classification is included as a baseline. The best result is highlighted in bold. Note, that the size of the dataset of \cifairC with \num{5} samples per class is the same as that of \cifairX and \stl with \num{50} samples.
}
\end{table*}

\section{Related Work}
\label{sec:related_work}

Learning deep neural network from limited amount of labeled data has been studied from various perspectives. 
Many methods have been proposed in the field of transfer learning and few-shot learning. 
Small data learning differs from both research areas since all networks are trained from scratch and only a small number of labeled training examples is used without any additional data.

Deep neural networks have thousands or millions of parameters that need to be trained. 
To avoid overfitting, various regularization techniques have been presented such as Dropout \cite{srivastavaDropoutSimpleWay2014} or BatchNormalization \cite{ioffeBatchNormalizationAccelerating2015}.
However, when training on small datasets these regularization mechanisms are not able to prevent overfitting and other ways of regularization are required.
\paragraph{Data Augmentation.}
\label{par:data_augmentation}

Data augmentation increases the size of the dataset by applying random transformations to the original data samples generating new synthetic training samples.
Standard transformations for image classification are random cropping of patches and horizontal or vertical flipping \cite{krizhevskyImageNetClassificationDeep2012}.
Cutout \cite{devriesImprovedRegularizationConvolutional2017} masks a square region in the image to improve the robustness of the networks. 
Similarly, Random Erasing \cite{zhongRandomErasingData2020} replaces a region in the image with random values.
Automatic augmentation methods for learning the augmentation strategy have been proposed in recent years combining multiple geometric and color transformations. 
AutoAugment \cite{cubukAutoAugmentLearningAugmentation2019} optimizes the parameters of the policies using a recurrent neural network via reinforcement learning. 
Since this process is computationally expensive and requires training data, \textcite{mullerTrivialAugmentTuningfreeStateoftheArt2021} propose TrivialAugment which is parameter-free. 
The augmentation method randomly samples a single policy per image that is applied with a uniformly sampled strength.

The performance of different network architectures in relation to the amount of training data has been analyzed by \textcite{brigatoCloseLookDeep2021}. 
Simple models achieve good results when little data is available, however, deeper networks catch up when augmentation is applied.
The authors suggest that further data generation and augmentation methods could further boost the performance.
\textcite{bornscheinSmallDataBig2020} analyze the generalization performance of deep neural networks depending on the size of the dataset empirically and found that large networks mostly outperform small networks even on small datasets.
\textcite{barzDeepLearningSmall2020} propose the use of the cosine loss function instead of the cross-entropy loss, or a combination of both, in the small data regime. 
While the cross-entropy loss moves the activations towards infinity, the cosine loss function includes an $l^2$ normalization as a regularizer.
\textcite{aroraHarnessingPowerInfinitely2020} explore the performance of convolutional neural tangent kernels (CNTK) compared to a ResNet trained on small amount of data. 
The combination of convolutional neural networks for feature learning and random forests as robust classifiers has been demonstrated in  \cite{reindersObjectRecognitionVery2018,ReiAck2019a}.
\textcite{gauthierParametricScatteringNetworks2021} present a parametric scattering transform to learn descriptive representations. 
The authors introduce a differentiable architecture that learns the parameters of the Morlet wavelet filters.

\paragraph{Mixing Augmentation.}
\label{par:mixing_augmentation}

While classic data augmentation processes one image at a time, there are several approaches that use multiple samples.
MixUp \cite{zhangMixupEmpiricalRisk2018} generates weighted combinations of random image pairs by linear interpolation between the images and targets.
CutMix \cite{yunCutMixRegularizationStrategy2019} replaces a rectangular region of one image with the content from another image.
Generative approaches for synthesizing new samples have been used in a variety of domains, such as video game levels \cite{Awiszus_Schubert_Rosenhahn_2020,SchuAwi2021} and object detection \cite{KluRei2018}.
\ac{GLICO} \cite{azuriGenerativeLatentImplicit2021} is a generative method for synthesizing new samples using spherical interpolation in the latent space. The method has a learnable latent representation for each training sample and optimizes the reconstruction by the generator. 
An additional classification loss tunes the semantic structure of the latent space. 
Finally, a method that also uses masks for data augmentation is SuperMix \cite{daboueiSuperMixSupervisingMixing2021}.
SuperMix introduces a student-teacher architecture to optimize the masks according to which the two images are mixed.

\begin{table*}[t]
\centering
\label{tab:results_aa_ta_main}
\begin{tabular}{lccS[table-format=2.2(2)]S[table-format=2.2(2)]S[table-format=2.2(2)]S[table-format=2.2(2)]S[table-format=2.2(2)]S[table-format=2.2(2)]S[table-format=2.2(2)]S[table-format=2.2(2)]}
\toprule
\multicolumn{2}{l}{Samples per Class} & {} & {5} & {10} & {20} & {30} & {50} & {100} \\
{Method} & {AA} & {TA} & {} & {} & {} & {} & {} & {} \\
\midrule
AutoAugment & \tick & \cross & \SI{21.39\pm0.95}{} & \SI{29.56\pm0.68}{} & \SI{40.81\pm0.35}{} & \SI{47.58\pm0.56}{} & \SI{55.01\pm0.24}{} & \SI{63.69\pm0.42}{} \\
TrivialAugment & \cross & \tick & \SI{23.85\pm0.60}{} & \SI{32.48\pm0.34}{} & \SI{44.13\pm0.16}{} & \SI{50.17\pm0.26}{} & \SI{56.27\pm0.19}{} & \SI{64.02\pm0.18}{} \\
ChimeraMix+Grid & \cross & \cross & \SI{20.24\pm0.12}{} & \SI{31.62\pm0.82}{} & \SI{41.80\pm0.52}{} & \SI{48.10\pm0.71}{} & \SI{54.67\pm1.01}{} & \SI{62.13\pm0.27}{} \\
ChimeraMix+Grid & \tick & \cross & \SI{25.24\pm1.02}{} & \SI{34.60\pm0.47}{} & \SI{45.16\pm0.38}{} & \SI{51.00\pm0.87}{} & \SI{57.74\pm0.51}{} & \SI{64.19\pm0.68}{} \\
ChimeraMix+Grid & \cross & \tick & \SI{25.69\pm0.37}{} & \SI{34.67\pm0.51}{} & \SI{45.78\pm0.10}{} & \SI{51.81\pm0.11}{} & \SI{57.80\pm0.62}{} & \SI{64.21\pm0.37}{} \\
ChimeraMix+Seg & \cross & \cross & \SI{21.09\pm0.47}{} & \SI{32.72\pm0.60}{} & \SI{43.23\pm0.38}{} & \SI{48.83\pm0.72}{} & \SI{55.79\pm0.21}{} & \SI{62.96\pm0.77}{} \\
ChimeraMix+Seg & \tick & \cross & \SI{25.16\pm0.37}{} & \SI{35.02\pm0.55}{} & \SI{45.27\pm0.09}{} & \SI{51.25\pm0.67}{} & \SI{57.86\pm0.41}{} & \SI{64.39\pm0.43}{} \\
ChimeraMix+Seg & \cross & \tick & \bfseries \SI{26.36\pm0.17}{} & \bfseries \SI{36.02\pm0.22}{} & \bfseries \SI{46.61\pm0.38}{} & \bfseries \SI{52.74\pm0.20}{} & \bfseries \SI{58.90\pm0.64}{} & \bfseries \SI{64.79\pm0.06}{} \\
\bottomrule
\end{tabular}
\caption{Accuracy of \methodname with AutoAugment (AA) and TrivialAugment (TA) on \cifairC. \cifairX and \stl are shown in the supplementary material.} %
\end{table*}

\section{Method}
\label{sec:method}

In this work, we present \methodname, a novel method for learning deep neural networks with a small amount of data. 
\methodname is a generative approach that learns to create new compositions of image pairs from the same class by mixing their features guided by a mask.

\subsection{Mixing Generator}

\methodname's general architecture is shown in \cref{fig:architecture}.
Given a labeled image dataset, in each iteration a pair of images $x_1, x_2 \in \mathbb{R}^{H \times W \times C}$ from the same class is sampled, where $H \times W$ is the image size and $C$ the number of color channels.
\methodname consists of an encoder and a decoder module.
The encoder $f$ extracts features $e_i = f(x_i) \in \mathbb{R}^{H' \times W' \times C'}$ with a downsampled size of $H' \times W'$ and $C'$ dimensions. 
At each spatial location, a mask $m \in \{0, 1\}^{H' \times W'}$ selects the feature vector from one of both images.
Finally, the decoder $g$ generates new image compositions $\hat{x}$ based on the mixed features, i.e.,
\begin{equation}
    \hat{x} = g \left( e_1 \odot m + e_2 \odot (1 - m) \right).
    \label{eq:generation_process}
\end{equation}

\paragraph{Training.}

To generate valid compositions, we introduce a discriminator $D$ that tries to distinguish real and generated images performing a binary classification.
During training, the discriminator is optimized by minimizing the mean squared error $\mathcal{L}_{\text{D,disc}}$ between the predicted label and actual label.
The generator, on the other hand, learns to create realistic image compositions by minimizing the mean squared error between the discriminator's prediction of the generated image and the real image class (denoted as $\mathcal{L}_{\text{G,disc}}$), i.e., the discriminator cannot identify if the image is real or generated. 
The discriminator and generator are trained alternately.
Additionally, the generator is guided by reconstructing the original images. 
When the mask $m$ is set to zeros, the image $x_1$ should be reconstructed, and vice versa for $x_2$ when $m$ consists of ones. 
The reconstruction loss minimizes the similarity between the generator's output and the corresponding input image and is defined as 
\begin{equation}
    \mathcal{L}_{\text{rec}} = ||\hat{x}_1 - x_1||^2 + ||\hat{x}_2 - x_2||^2.
\end{equation}
It is common to improve the visual appearance of generated images by using a perceptual loss that is computed using the features of a VGG \cite{simonyanVeryDeepConvolutional2015} network.
However, as we are in the small data setting, we have to resort to another method to match the appearance of our outputs and the input images.
Our loss $\mathcal{L}_{\text{per}}$ uses a laplacian pyramid of image-output pairs and computes the $l^1$-distance between them \cite{dentonDeepGenerativeImage2015}.
Thus, we minimize the following generator loss 
\begin{equation}
    \mathcal{L}_{\text{G}} = \alpha_{\text{rec}} \mathcal{L}_{\text{rec}} + \alpha_{\text{per}} \mathcal{L}_{\text{per}} + \alpha_{\text{disc}} \mathcal{L}_{\text{G,disc}},
    \label{eq:loss}
\end{equation}
where $\alpha_{\text{rec}}$, $\alpha_{\text{per}}$, and $\alpha_{\text{disc}}$ are weightings.
The exact parameters can be found in the supplementary material. %

\paragraph{Classification.}
\label{par:classifier}

After \methodname has been trained on a given dataset, it is used to augment the dataset and train an arbitrary classifier. 
The generative model enables the composition of new images by sampling combinations of image pairs and different masks. 
In this way, we are able to greatly increase the dataset. 
For training the classifier, each batch is replaced with a probability of $50\%$ with compositions that are generated by \methodname.

\subsection{Masks}
\label{subsec:masks}

\begin{figure}[!h]
     \centering
     \begin{subfigure}[b]{0.32\linewidth}
         \centering
         \includegraphics[width=\linewidth]{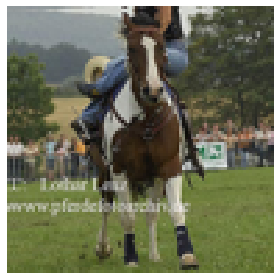}
         \caption{Original image}
         \label{fig:stl10_orig}
     \end{subfigure}
     \hfill
     \begin{subfigure}[b]{0.32\linewidth}
         \centering
         \includegraphics[width=\linewidth]{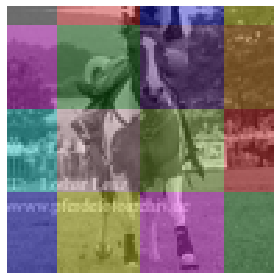}
         \caption{Grid}
         \label{fig:stl10_grid_mask}
     \end{subfigure}
     \hfill
     \begin{subfigure}[b]{0.32\linewidth}
         \centering
         \includegraphics[width=\linewidth]{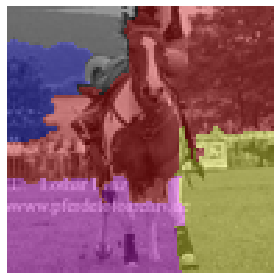}
         \caption{Segmentation}
         \label{fig:stl10_seg_mask}
     \end{subfigure}
    \caption{Example image from \stl as well as a visualisation of a grid and segmentation mask. For each region, we sample from a binomial distribution. The masks guide the feature mixing in the generator.}
    \label{fig:stl10_mix_masks}
\end{figure}

We evaluate two different ways of generating the mask that guides the mixing of the image features in \methodname.
See \cref{fig:stl10_mix_masks} for an example.

\paragraph{Grid.}
\label{par:grid}

The masks can be sampled as a grid.
For that, we sample $m$ from a binomial distribution $\mathcal{B}$ and interpolate it to the size of the feature tensor $H' \times W'$.
The dimension of $m$, i.e., the size of the blocks, is a hyperparameter and depends on the size of the structural components in the image. 
By generating a grid structure, blocks of features are mixed in the generator while maintaining the local structure of the instances.

\paragraph{Segmentation.}
\label{par:segmentation_masks}

However, we can improve on this heuristic by taking the image gradients into account.
Classic algorithms such as Felzenszwalb segmentation \cite{felzenszwalbEfficientGraphBasedImage2004} enable us to mix the features of contiguous regions in the images.
The regions are computed from a graph representation of the image that is based on pixel similarities.
For \methodname, any segmentation algorithm might be used, as the segmentation masks can be precomputed and do not cause significantly more computational cost.
We choose Felzenszwalb segmentation as it has only few hyperparameters\footnote{We chose the initial parameters according to \textcite{henaffEfficientVisualPretraining2021} and adjusted them to the smaller image sizes.}.
Using the segmentation masks, we uniformly sample a segmented region from the image $x_1$, create the corresponding binary mask, and downsample it to the size of the feature $e_1$.
This leads to more expressive samples.
We call this variant \emph{\methodnameseg}.

\section{Experiments}
\label{sec:experiments}

We evaluate the performance of \methodname and current state-of-the-art methods on benchmark datasets and analyze their performance on different dataset sizes.
To study the visual quality of the images that \methodname generates, we compare their \ac{FID} with that of \ac{GLICO} while following the methodology of \textcite{parmarBuggyResizingLibraries2021}.
Due to the generative nature of \methodname, we can combine it with standard augmentation methods such as AutoAugment and TrivialAugment.
Finally, we perform a sensitivity analysis of our approach in the supplementary material and an ablation regarding \methodname's generator. %

\begin{figure*}[!ht]
    \centering
    \includegraphics[width=\linewidth]{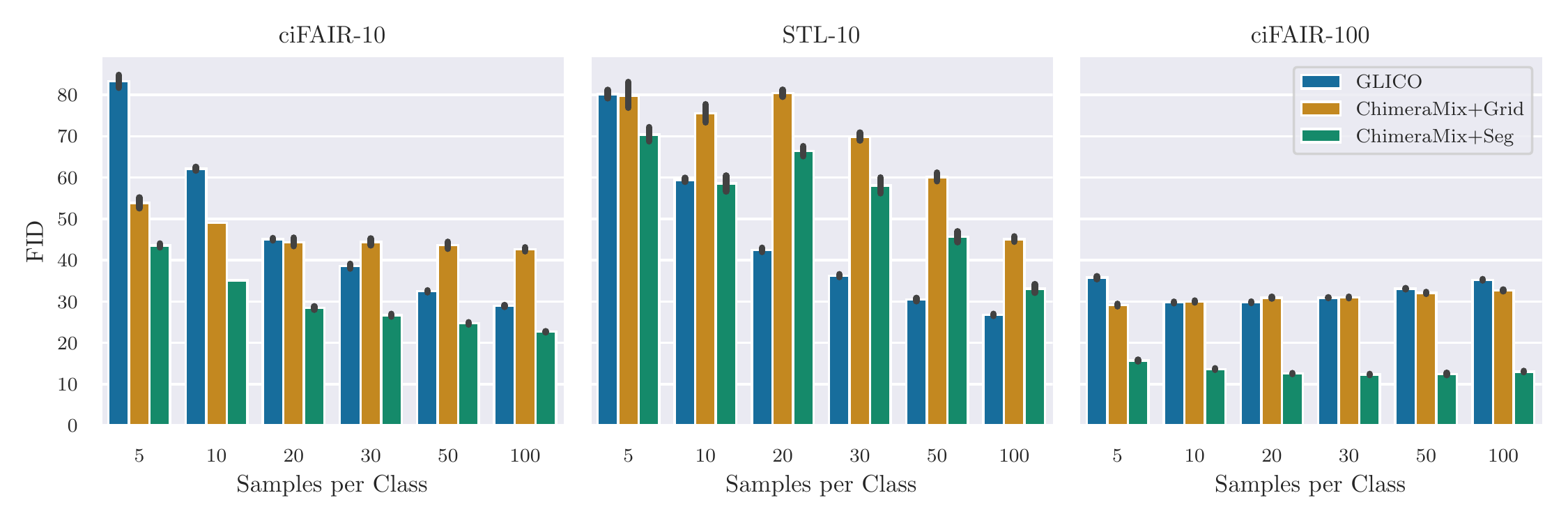}
    \caption{\acl{FID}$(\downarrow)$ of \ac{GLICO}, \methodnamegrid, and \methodnameseg on \cifairX, \stl, and \cifairC.}
    \label{fig:fid}
\end{figure*}

\subsection{Datasets}

For our experiments, we choose the \cifairX and \cifairC \cite{barzWeTrainTest2020a} datasets.
Both contain \num{50000} and \num{10000} images of size $32 \times 32$ in their training and test set and have been extensively used in computer vision research\footnote{The \cifairX and \cifairC datasets have the same training set as their CIFAR \cite{krizhevskyLearningMultipleLayers2009} counterparts, but provide a cleaned test set.}.
Lastly, we evaluate our method on \stl \cite{coatesAnalysisSingleLayerNetworks2011} which is a slightly more complex dataset consisting of \num{5000} training and \num{8000} test images of size $96 \times 96$ from $10$ categories.
Even though these datasets already are quite small compared to other contemporary datasets such as ImageNet \cite{russakovskyImageNetLargeScale2015}, we further subsample the labels per class to evaluate the algorithms in the small data regime.
For all datasets we sample the instances uniformly from each class.

\subsection{Experimental Setup}

We use the same architecture for all methods.
On \cifairX and \cifairC we train a WideResNet-16-8 and on \stl we use a ResNet-50 due to the larger image size.
We use SGD with momentum and a cosine-annealing learning rate schedule with weight decay.
The hyperparameters can be found in the supplementary material. %
All experiments are repeated five times on different dataset splits.

\subsection{Comparison with State-of-the-Art}
\label{subsec:comparison_to_sota}

We compare \methodname on all datasets against several methods such as Cutout \cite{devriesImprovedRegularizationConvolutional2017}, Random Erasing \cite{zhongRandomErasingData2020}, Cosine loss \cite{barzDeepLearningSmall2020},  MixUp \cite{zhangMixupEmpiricalRisk2018}, and \ac{GLICO} \cite{azuriGenerativeLatentImplicit2021}.
On \cifairX and \cifairC we compare against Parametric Scattering Networks \cite{gauthierParametricScatteringNetworks2021} which showed state-of-the-art in the small data regime on \cifairX \footnote{Parametric Scattering Networks uses a modified WideResNet that has more units in the first layers. Since the performance with a standard WideResNet degrades, we evaluate their method with the modified architecture.}.
Additionally, we include a comparison of SuperMix \cite{daboueiSuperMixSupervisingMixing2021} on \cifairC. 
For all datasets, a standard classification network is evaluated as a baseline.

The results are shown in \cref{tab:results_cifair10_stl10} for different numbers of samples per class.
On \cifairX with $5$ samples per class, for example, where only $50$ training examples are available, the current best performing method achieves $33.41\%$. 
\methodnamegrid and \methodnameseg reach a test performance of $36.94\%$ and $37.31\%$, respectively.
Overall, the results demonstrate that \methodnamegrid and \methodnameseg generate image compositions that improve the training and lead to a higher accuracy, especially in the small data regime.

\subsection{Comparison with TrivialAugment and AutoAugment}

In the next experiment, we investigate whether \methodname can be combined with known methods for automatic data augmentation such as AutoAugment \cite{cubukAutoAugmentLearningAugmentation2019} and TrivialAugment \cite{mullerTrivialAugmentTuningfreeStateoftheArt2021}.
It should be noted, that the policies of AutoAugment are optimized on the entire dataset.
The performance of AutoAugment, TrivialAugment, \methodname, and combinations on \cifairC are shown in \cref{tab:results_aa_ta_main}. 
Evaluations on the other datasets are presented in the supplementary material. %

The results show that TrivialAugment achieves a higher accuracy than AutoAugment on \cifairC, and vice versa on \cifairX and \stl.
\methodname, without any severe data augmentation, is already able to reach a similar performance as AutoAugment.
The combination of \methodname and AutoAugment or TrivialAugment significantly increases the performance. 
On \cifairC with 5 samples per class, \methodnameseg achieves $26.48\%$ in combination with TrivialAugment compared to $23.85\%$ without \methodnameseg.

\begin{table*}[h]
\label{tab:ablation_generator_direct}
\begin{tabular}{llccS[table-format=2.2(2)]S[table-format=2.2(2)]S[table-format=2.2(2)]S[table-format=2.2(2)]S[table-format=2.2(2)]S[table-format=2.2(2)]}
\toprule
\multicolumn{2}{l}{Samples per Class} & {5} & {10} & {20} & {30} & {50} & {100} \\
{Dataset} & {Method} & {} & {} & {} & {} & {} & {} \\
\midrule
\multirow[c]{4}{*}{ciFAIR-10} & GridMix & \SI{29.97\pm1.17}{} & \SI{39.90\pm1.24}{} & \SI{48.60\pm3.18}{} & \SI{54.99\pm2.49}{} & \SI{61.12\pm1.59}{} & \SI{72.41\pm0.69}{} \\
 & SegMix & \SI{32.00\pm1.22}{} & \SI{42.18\pm1.36}{} & \SI{52.56\pm2.43}{} & \SI{57.90\pm0.49}{} & \SI{64.61\pm0.94}{} & \SI{73.96\pm0.36}{} \\
 & ChimeraMix+Grid & \SI{36.94\pm2.63}{} & \SI{45.57\pm2.11}{} & \SI{53.67\pm2.84}{} & \SI{59.66\pm1.35}{} & \SI{65.42\pm0.83}{} & \SI{73.76\pm0.30}{} \\
 & ChimeraMix+Seg & \bfseries \SI{37.31\pm2.57}{} & \bfseries \SI{47.60\pm1.81}{} & \bfseries \SI{56.21\pm1.77}{} & \bfseries \SI{60.92\pm0.62}{} & \bfseries \SI{67.30\pm1.21}{} & \bfseries \SI{74.96\pm0.21}{} \\
\midrule
\multirow[c]{4}{*}{STL-10} & GridMix & \SI{28.98\pm1.49}{} & \SI{31.21\pm1.52}{} & \SI{37.08\pm1.09}{} & \SI{42.14\pm1.52}{} & \SI{49.33\pm0.88}{} & \SI{56.92\pm0.51}{} \\
 & SegMix & \SI{29.25\pm0.40}{} & \SI{32.84\pm0.63}{} & \SI{37.80\pm1.91}{} & \SI{43.69\pm0.84}{} & \SI{50.14\pm0.84}{} & \SI{58.60\pm0.57}{} \\
 & ChimeraMix+Grid & \bfseries \SI{32.18\pm0.90}{} & \SI{37.01\pm0.84}{} & \SI{43.19\pm1.03}{} & \SI{48.93\pm1.34}{} & \SI{52.81\pm1.45}{} & \SI{60.04\pm0.27}{} \\
 & ChimeraMix+Seg & \SI{31.37\pm1.72}{} & \bfseries \SI{37.05\pm1.09}{} & \bfseries \SI{44.74\pm0.60}{} & \bfseries \SI{49.58\pm0.49}{} & \bfseries \SI{55.06\pm1.11}{} & \bfseries \SI{60.44\pm0.71}{} \\
\midrule
\multirow[c]{4}{*}{ciFAIR-100} & GridMix & \SI{17.98\pm0.23}{} & \SI{27.78\pm0.46}{} & \SI{38.92\pm0.05}{} & \SI{45.16\pm1.05}{} & \SI{52.97\pm0.30}{} & \SI{61.37\pm0.26}{} \\
 & SegMix & \SI{19.36\pm0.82}{} & \SI{29.62\pm0.22}{} & \SI{41.00\pm0.42}{} & \SI{47.50\pm0.38}{} & \SI{54.62\pm0.15}{} & \SI{62.43\pm0.38}{} \\
 & ChimeraMix+Grid & \SI{20.24\pm0.12}{} & \SI{31.62\pm0.82}{} & \SI{41.80\pm0.52}{} & \SI{48.10\pm0.71}{} & \SI{54.67\pm1.01}{} & \SI{62.13\pm0.27}{} \\
 & ChimeraMix+Seg & \bfseries \SI{21.09\pm0.47}{} & \bfseries \SI{32.72\pm0.60}{} & \bfseries \SI{43.23\pm0.38}{} & \bfseries \SI{48.83\pm0.72}{} & \bfseries \SI{55.79\pm0.21}{} & \bfseries \SI{62.96\pm0.77}{} \\
\bottomrule
\end{tabular}
\caption{Analysis of the generator's impact. GridMix and SegMix directly mix the images without the generator of \methodname. The study shows that mixing the feature via the proposed generator (\methodnamegrid and \methodnameseg) is able to learn the generation of new image compositions and achieves a significantly improved performance.
}
\end{table*}

\subsection{Analysis of Generated Samples}
\label{subsec:analysis_of_generated_samples}

We conduct an analysis of the generated samples similar to \ac{GLICO} \cite{azuriGenerativeLatentImplicit2021} since both are generative approaches.
The samples of \methodname should be valid compositions of the given image pairs and thus have similar visual features.
\cref{fig:stl10_samples} shows qualitative example outputs of \methodname on \stl.
The generator has learned to combine the two images of a cat in sensible ways, e.g., by placing the head of one cat on the body of the other cat.
If these features are diverse enough and close to the original full training set, the classifier should be able to achieve a similar validation accuracy.
We compare the features of our generated datasets by using an Inception-v3 \cite{szegedyGoingDeeperConvolutions2015} network that was pretrained on ImageNet and compute the \ac{FID} \cite{heuselGANsTrainedTwo2017} to the original full training set\footnote{We use the code provided by \textcite{parmarBuggyResizingLibraries2021}.}.
The \ac{FID} is defined as 
\begin{equation}
    \text{FID} = ||\mu - \mu_{\omega}||_2^2 + \text{tr}\left(\Sigma + \Sigma_{\omega} - 2\left(\Sigma^{\frac{1}{2}} \Sigma_{\omega} \Sigma^{\frac{1}{2}}\right)^{\frac{1}{2}}\right)
\end{equation}
comparing the gaussian approximations of the activations between the two datasets.
$\mu_{\omega}$ and $\Sigma_{\omega}$ are the mean and covariance of the full training set.

The results are shown in \cref{fig:fid} and indicate that \methodname is able to produce samples that are close to the full training set.
Compared to the \ac{GLICO} baseline, except for \stl, we achieve better \ac{FID} scores.
A visualization of the representation of the augmented dataset is shown in the supplementary material. %

\subsection{Ablation Study}
\label{subsec:ablation_study}

To analyze the effect of our generator-discriminator architecture, we compare \methodname with two versions \emph{GridMix} and \emph{SegMix} that mix the images directly using our two methods to produce masks.
Given two images and a mask, we are already able to generate compositions and train a classifier on them.
Note, that GridMix is similar to CutMix by \textcite{yunCutMixRegularizationStrategy2019} except that the patches consist of multiple possibly overlapping rectangles.
The results are shown in \cref{tab:ablation_generator_direct}.
It is apparent that the generator of \methodname improves the quality of the mixed images and leads to a significant improvement. 
On \cifairX with $10$ samples per class, for example, \methodnamegrid increases the accuracy from $39.90\%$ to $45.57\%$ and \methodnameseg from $42.18\%$ to $47.60\%$.

\section{Conclusion}
\label{sec:conclusion}

In this work, we presented \methodname, a novel method to improve the performance of image classifiers given only a handful of samples per class.
Our generative approach mixes the features of two images from the same class using a binary mask and learns to generate new samples that are compositions of the given images.
The mask is either sampled from a binomial distribution or generated from the segmented areas in both images.
The experiments show that \methodname is able to generate new image compositions that significantly improve the classification. 
We demonstrate state-of-the-art classification performance on several datasets and investigate the distribution of the compositions that our method generates using the \acl{FID}.
In future work, our method can be extended to mix multiple images.
Additionally, the masks can be refined by integrating techniques such as SuperMix after they have been adjusted to the small data setting.

\section*{Acknowledgments}
This work was supported by the Federal Ministry of Education and
Research (BMBF), Germany, under the project LeibnizKILabor (grant no.
01DD20003), the Center for Digital Innovations (ZDIN),
 and the Deutsche Forschungsgemeinschaft  (DFG)  under  Germany’s  Excellence  Strategy  within  the  Cluster of Excellence PhoenixD (EXC 2122).
 
\bibliographystyle{named}
{\small
\bibliography{main}
}

\newpage

\begin{appendices}
\onecolumn

\begin{center}
\LARGE
\textbf{Supplementary Material} \\
\vspace{0.25cm}
\textbf{\methodname: Image Classification on Small Datasets via Masked Feature Mixing}
\vspace{1cm}
\end{center}

\section{Evaluating the Inception Representations}
\label{sec:appendix_inception_representations}

\begin{figure}[!hb]
    \centering
    \includegraphics[width=\linewidth]{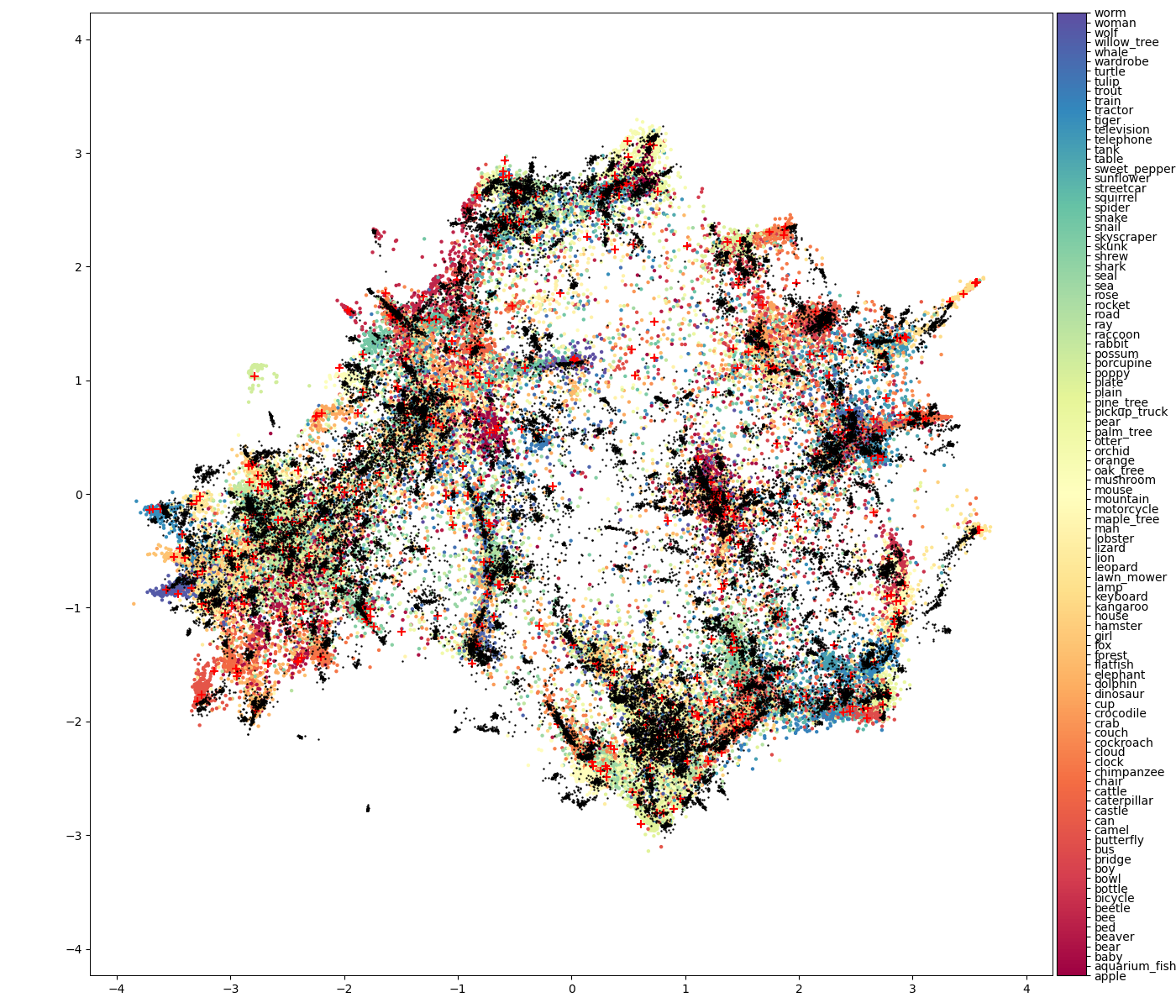}
    \caption{Embedding using \ac{MDE} of Inception representations on \cifairC using 5 samples per class. All samples from the entire dataset are visualized as colored circles. 
    The black dots are generated samples and the red crosses depict the subsampled training data of \methodnameseg.}
    \label{fig:mde_cifar100_5}
\end{figure}

Additionally to the \ac{FID}, we evaluate whether \methodname is able to generate meaningful samples by analysing the Inception-v3 embeddings via the \ac{MDE} \citeSM{agrawalMinimumDistortionEmbedding2021} method in \cref{fig:mde_cifar100_5}.

\newpage
\section{Results for TrivialAugment and AutoAugment}
\label{sec:appendix_autoaugment}

AutoAugment and TrivialAugment have been successful in producing state-of-the-art image classification results by tuning a set of standard augmentations on several datasets using a Reinforcement Learning agent.
The results of AutoAugment, TrivialAugment, and combinations with \methodname are shown in \cref{tab:results_aa_ta_supplementary}.

\begin{table}[h]
\label{tab:results_aa_ta_supplementary}
\begin{tabular}{llccS[table-format=2.2(2)]S[table-format=2.2(2)]S[table-format=2.2(2)]S[table-format=2.2(2)]S[table-format=2.2(2)]S[table-format=2.2(2)]S[table-format=2.2(2)]}
\toprule
\multicolumn{4}{l}{Samples per Class} & {5} & {10} & {30} & {50} & {100} \\
{Dataset} & {Method} & {AA} & {TA} & {} & {} & {} & {} & {} \\
\midrule
\multirow[c]{8}{*}{ciFAIR-10} & AutoAugment & \tick & \cross & \SI{35.64\pm3.23}{} & \SI{44.43\pm2.19}{} & \SI{60.80\pm0.75}{} & \SI{67.18\pm0.99}{} & \SI{74.86\pm0.40}{} \\
 & TrivialAugment & \cross & \tick & \SI{31.55\pm3.77}{} & \SI{41.87\pm1.56}{} & \SI{58.60\pm1.08}{} & \SI{66.30\pm0.94}{} & \SI{75.54\pm0.50}{} \\
 & ChimeraMix+Grid & \cross & \cross & \SI{36.94\pm2.63}{} & \SI{45.57\pm2.11}{} & \SI{59.66\pm1.35}{} & \SI{65.42\pm0.83}{} & \SI{73.76\pm0.30}{} \\
 & ChimeraMix+Grid & \tick & \cross & \SI{41.28\pm1.62}{} & \SI{49.02\pm1.41}{} & \SI{64.13\pm0.29}{} & \SI{69.90\pm0.32}{} & \SI{76.91\pm0.49}{} \\
 & ChimeraMix+Grid & \cross & \tick & \SI{35.86\pm3.11}{} & \SI{45.32\pm1.96}{} & \SI{61.69\pm0.75}{} & \SI{69.04\pm0.10}{} & \SI{76.88\pm0.67}{} \\
 & ChimeraMix+Seg & \cross & \cross & \SI{37.31\pm2.57}{} & \SI{47.60\pm1.81}{} & \SI{60.92\pm0.62}{} & \SI{67.30\pm1.21}{} & \SI{74.96\pm0.21}{} \\
 & ChimeraMix+Seg & \tick & \cross & \bfseries \SI{42.16\pm1.00}{} & \bfseries \SI{49.75\pm1.55}{} & \bfseries \SI{65.28\pm0.32}{} & \SI{70.09\pm0.72}{} & \SI{76.76\pm0.35}{} \\
 & ChimeraMix+Seg & \cross & \tick & \SI{36.74\pm3.55}{} & \SI{46.58\pm2.15}{} & \SI{63.21\pm0.48}{} & \bfseries \SI{70.24\pm0.85}{} & \bfseries \SI{77.79\pm0.46}{} \\
\midrule
\multirow[c]{8}{*}{STL-10} & AutoAugment & \tick & \cross & \SI{32.05\pm0.93}{} & \SI{37.65\pm2.26}{} & \SI{49.77\pm1.09}{} & \SI{53.84\pm0.96}{} & \SI{59.55\pm0.96}{} \\
 & TrivialAugment & \cross & \tick & \SI{30.91\pm1.98}{} & \SI{35.87\pm1.57}{} & \SI{47.67\pm0.56}{} & \SI{53.50\pm1.89}{} & \SI{61.04\pm0.70}{} \\
 & ChimeraMix+Grid & \cross & \cross & \SI{32.18\pm0.90}{} & \SI{37.01\pm0.84}{} & \SI{48.93\pm1.34}{} & \SI{52.81\pm1.45}{} & \SI{60.04\pm0.27}{} \\
 & ChimeraMix+Grid & \tick & \cross & \bfseries \SI{37.54\pm1.74}{} & \SI{43.12\pm0.79}{} & \SI{53.75\pm1.38}{} & \bfseries \SI{57.76\pm1.46}{} & \SI{61.86\pm1.06}{} \\
 & ChimeraMix+Grid & \cross & \tick & \SI{34.88\pm1.97}{} & \SI{41.02\pm0.60}{} & \SI{51.86\pm1.02}{} & \SI{56.61\pm0.76}{} & \SI{62.43\pm0.11}{} \\
 & ChimeraMix+Seg & \cross & \cross & \SI{31.37\pm1.72}{} & \SI{37.05\pm1.09}{} & \SI{49.58\pm0.49}{} & \SI{55.06\pm1.11}{} & \SI{60.44\pm0.71}{} \\
 & ChimeraMix+Seg & \tick & \cross & \SI{36.71\pm1.46}{} & \bfseries \SI{43.88\pm0.73}{} & \bfseries \SI{54.90\pm1.08}{} & \SI{56.41\pm2.13}{} & \SI{60.98\pm0.96}{} \\
 & ChimeraMix+Seg & \cross & \tick & \SI{34.53\pm2.01}{} & \SI{41.08\pm0.44}{} & \SI{52.03\pm1.80}{} & \SI{55.66\pm0.72}{} & \bfseries \SI{63.83\pm0.52}{} \\
\bottomrule
\end{tabular}
\caption{Accuracy of \methodname with AutoAugment (AA) and TrivialAugment (TA) on \cifairX and \stl.}
\end{table}

\section{Hyperparameters}
\label{sec:appendix_hyperparameters}

\subsection{\methodnamegrid and \methodnameseg}

We weight the reconstruction loss by a factor of $\alpha_{\text{rec}}=1000$ and set $\alpha_{\text{per}} = \alpha_{\text{disc}} = 1$.
Due to the varying number of samples in the datasets, for \cifairX and \cifairC we repeat the data by a factor of $\lfloor\frac{500}{\text{Samples per Class}}\rfloor$.
For \stl, we decrease this factor to $\lfloor\frac{120}{\text{Samples per Class}}\rfloor$ to account for the smaller batch size and thus, higher number of gradient steps.

\subsubsection{Generator}

The generator architecture is based on the CycleGAN generator \citeSM{zhuUnpairedImagetoImageTranslation2017}. It consists of a downsampling module, a varying number of residual blocks, and an upsampling module. The feature mixing point defines the partition into encoder $f$ and generator $g$.
\cref{tab:appendix_hps_generator_common} lists the common hyperparameters to train the generator.

\begin{table}[h]
    \centering
    \begin{tabular}{c|c}
        Epochs & 200 \\
        Optimizer & Adam, $\beta$=[0.5, 0.999] \\
        Learning Rate Schedule & Stepwise Decay of \num{0.2}@[60, 120, 160] \\
        Initial Learning Rate & \num{0.0002} \\
        Weight Decay & \num{0.0005} \\
        Residual Blocks & 4 \\
        Feature Mixing after Block & 2 \\
        Mixing Mask Size & 4
    \end{tabular}
    \caption{Common hyperparameters of the \methodname generator for all datasets.}
    \label{tab:appendix_hps_generator_common}
\end{table}

For \cifairX and \cifairC we use a batch size of 64 while for \stl the batch size is 8 due to the differently sized images.
To prevent artifacts with small images, we upsample the $32 \times 32$ images of \cifairX and \cifairC to a size of $64 \times 64$ for the generator.
For the classification, the images are downsampled to their original size.

\paragraph{Generator Architecture}

\begin{figure}[!h]
    \centering
    \includegraphics[width=0.8\linewidth]{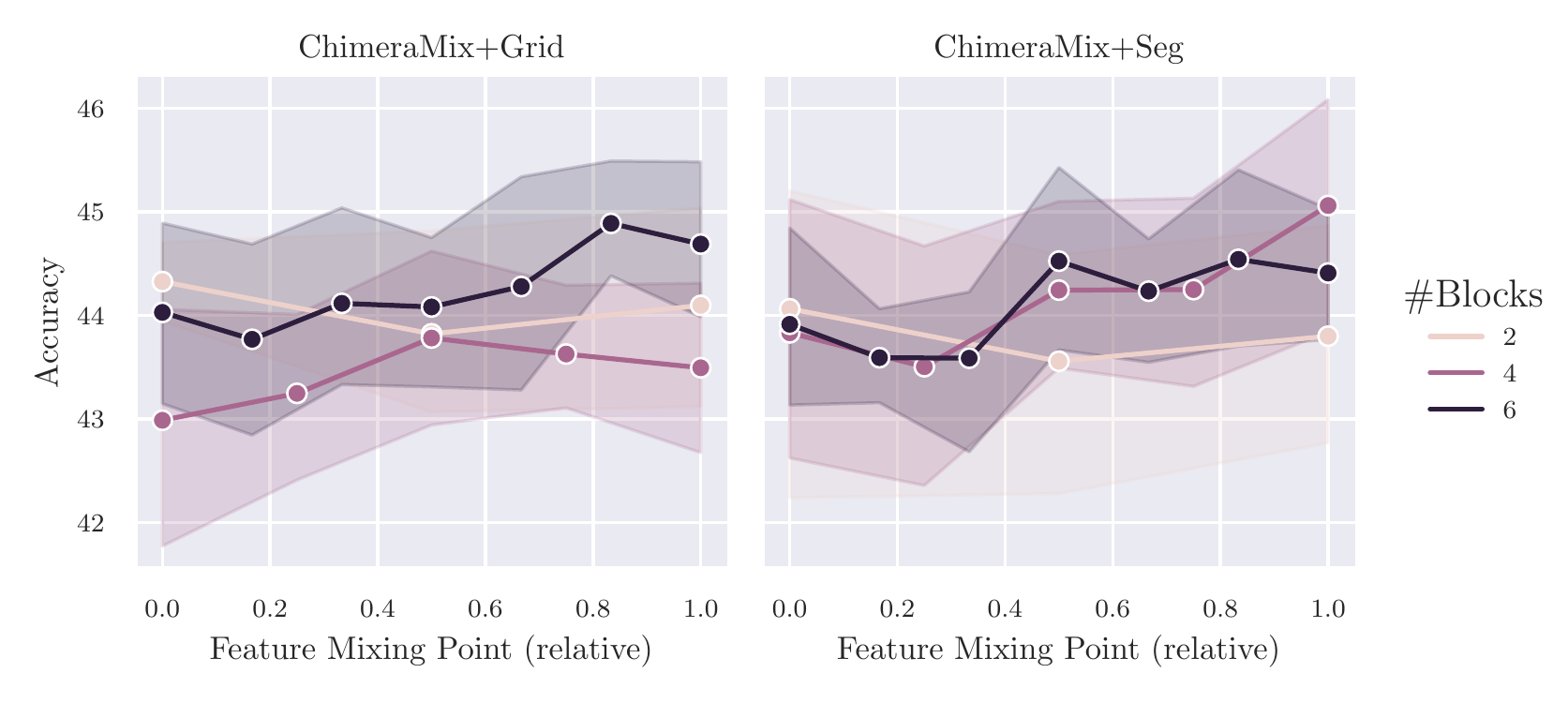}
    \caption{Analysis on \stl over the number of residual blocks and the point of feature mixing in \methodnamegrid and \methodnameseg. The x-axis is normalized to the number of blocks, i.e., the variant with four blocks at the relative feature mixing point $0.5$ mixes the features after two residual blocks.}
    \label{fig:ablation_generator_architecture}
\end{figure}

The architecture of \methodname has two main hyperparameters.
One is the number of residual blocks and the other is the point at which the features of the two provided images are mixed.
In \cref{fig:ablation_generator_architecture}, we perform a sensitivity analysis for these hyperparameters for \methodnamegrid and \methodnameseg on \stl with $20$ samples per class. Experiment shows the performance of generators with different numbers of residual blocks and feature mixing points. 
The results demonstrate that \methodnamegrid and \methodnameseg are robust to various generator architectures.

\paragraph{Grid Masks}

\begin{figure}[!h]
    \centering
    \includegraphics[width=0.4\linewidth]{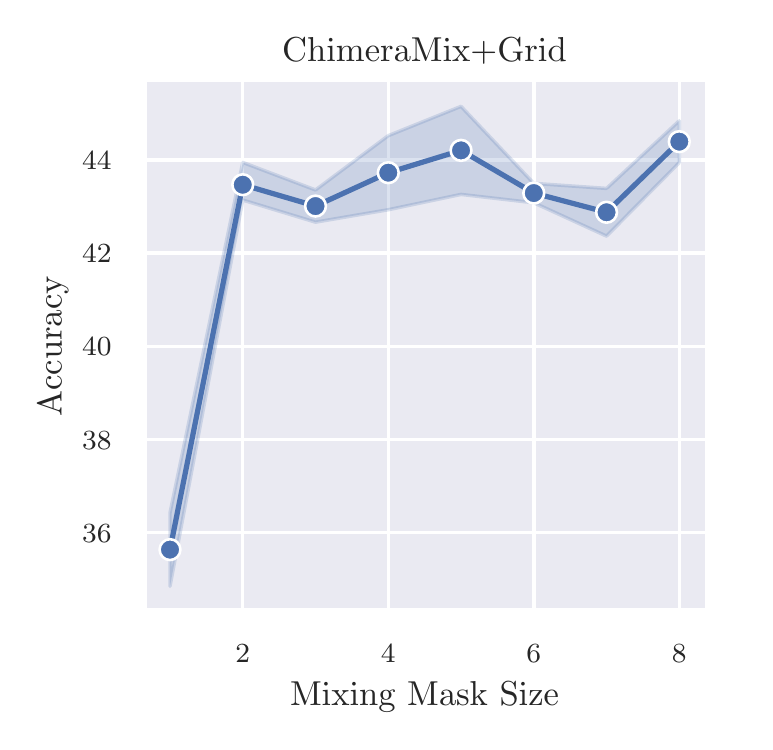}
    \caption{Analysis on \stl studying the mixing mask size, i.e., the size of the blocks in the sampled grid.}
    \label{fig:ablation_generator_masks}
\end{figure}

In the next experiment, we evaluate different mixing mask sizes for sampling the grid. They correspond to the number of blocks in the image. A higher number mixes smaller structures while a lower number focuses on larger components. The results are shown in \cref{fig:ablation_generator_masks} and demonstrate that \methodname is able to learn with different grid sizes.
When the mixing mask size is set to 1, the training is very similar to the baseline training with the only difference that the reconstructed images by the generator are used.

\subsubsection{Discriminator}

The discriminator architecture is similar to that of CycleGAN \citeSM{zhuUnpairedImagetoImageTranslation2017}.
It has 4 discriminator blocks consisting of convolutional layers with [\num{64}, \num{128}, \num{256}, \num{512}] channels that use instance normalization \citeSM{ulyanovInstanceNormalizationMissing2017} and Leaky ReLUs \citeSM{maasRectifierNonlinearitiesImprove2013}.
All convolutional filters have a kernel size of 4 and stride 1.
The final layer has a sigmoid activation function to discriminate each patch.

\subsubsection{Segmentation}

Felzenszwalb \citeSM{felzenszwalbEfficientGraphBasedImage2004} segmentation has two hyperparameters, the scale $s$ of the gaussian kernel that is used before performing the graph-based region merging and the minimal size of the regions $c$.
For \cifairX and \cifairC, we set both to \num{60}, while for \stl they are set to \num{400}.
We selected these parameters by following \citeSM{henaffEfficientVisualPretraining2021} who use $s=c=1000$ for their experiments on ImageNet.

\subsection{Classifier}

We use standard architectures for the classifiers to provide a fair comparison among all methods.
For \cifairX and \cifairC, we follow \citeSM{brigatoTuneItDon2021} and use a WideResNet-16-8.
For \stl, we use a ResNet-50 due to the larger image size.
The common hyperparameters are listed in \cref{tab:appendix_hps_classifier_common}.
As for the generator, we repeat the dataset according to the number of samples to keep the number of gradient steps similar.
The dataset specific hyperparameters (shown in Table \ref{tab:appendix_hps_classifier_cifair} and \ref{tab:appendix_hps_classifier_stl10}) are taken from \citeSM{brigatoTuneItDon2021}.

\begin{table}[h]
    \centering
    \begin{subtable}{.5\linewidth}
        \centering
        \begin{tabular}{c|c}
            Epochs & 200 \\
            Learning Rate Schedule & Cosine Annealing \\
            Optimizer & SGD \\
            Momentum & 0.9 \\
        \end{tabular}
        \caption{Common hyperparameters of the classifier for all datasets.}
        \label{tab:appendix_hps_classifier_common}
    \end{subtable}%
    \begin{subtable}{.5\linewidth}
        \centering
        \begin{tabular}{c|c}
            Batch Size & 10 \\
            Initial Learning Rate & 0.0046 \\
            Weight Decay & 0.0053 \\
            Dataset Repetitions & $\lfloor\frac{500}{\text{Samples per Class}}\rfloor$
        \end{tabular}
        \caption{Hyperparameters for the WideResNet-16-8 on \cifairX and \cifairC.}
        \label{tab:appendix_hps_classifier_cifair}
    \end{subtable} 
    \begin{subtable}{.5\linewidth}
        \centering
        \begin{tabular}{c|c}
            Batch Size & 16 \\
            Initial Learning Rate & 0.0074 \\
            Weight Decay & 0.00041 \\
            Dataset Repetitions & $\lfloor\frac{120}{\text{Samples per Class}}\rfloor$
        \end{tabular}
        \caption{Hyperparameters for the ResNet-50 on \stl.}
        \label{tab:appendix_hps_classifier_stl10}
    \end{subtable} 
\end{table}

\bibliographystyleSM{named}
\bibliographySM{main}

\end{appendices}

\end{document}